\documentclass{article} %
\usepackage{iclr2026_conference,times}

\usepackage{amsmath,amsfonts,bm}

\def\eqref#1{equation~\ref{#1}}

\def\1{\bm{1}}

\DeclareMathAlphabet{\mathsfit}{\encodingdefault}{\sfdefault}{m}{sl}
\SetMathAlphabet{\mathsfit}{bold}{\encodingdefault}{\sfdefault}{bx}{n}

\usepackage{hyperref}
\usepackage{url}
\usepackage{todonotes}
\usepackage{cleveref}
\usepackage{booktabs}
\usepackage{multirow}
\usepackage{subcaption}
\usepackage{listings}
\usepackage{xcolor}
\usepackage{enumitem}
\newcommand{\emoji}[1]{\raisebox{-0.15ex}{\includegraphics[height=1em]{#1}}}
\usepackage{tikz}
\usetikzlibrary{arrows.meta,positioning,calc}

\iclrfinalcopy

\title{On Randomness in Agentic Evals}

\author{Bjarni Haukur Bjarnason\thanks{Equal contribution.}~, André Silva\footnotemark[1]~, Martin Monperrus \\
KTH Royal Institute of Technology \\
Stockholm, Sweden \\
\texttt{\{bhbj, andreans, monperrus\}@kth.se}
}

\newcommand{\numTrajectories}{60,000}
\newcommand{\numTokens}{25.58B}
\newcommand{\numToolCalls}{1.88M}

\newcommand{\swebenchverified}{\texttt{SWE-Bench-Verified}}

\begin{document}

\maketitle

\begin{abstract}
Agentic systems are evaluated on benchmarks where agents interact with environments to solve tasks.
Most papers report a \texttt{pass@1} score computed from a single run per task, assuming this gives a reliable performance estimate.
We test this assumption by collecting \numTrajectories{} agentic trajectories on \swebenchverified{}, spanning three models and two scaffolds.
We find substantial variance: single-run \texttt{pass@1} estimates vary by 2.2 to 6.0 percentage points depending on which run is selected, with standard deviations exceeding 1.5 percentage points even at temperature 0.
This variance has critical implications: reported improvements of 2--3 percentage points may reflect evaluation noise rather than genuine algorithmic progress.
Through token-level analysis, we show that trajectories diverge early, often within the first few percent of tokens, and that these small differences cascade into different solution strategies.
To enable reliable evaluation of agentic systems, we recommend three concrete practices: (1) estimate \texttt{pass@1} from multiple independent runs per task, especially when measuring small improvements, (2) use statistical power analysis to determine the number of runs needed to detect expected effect sizes, and (3) consider metrics like \texttt{pass@k} (optimistic bound) and \texttt{pass\^{}k} (pessimistic bound) with $k>1$ to better characterize the full performance envelope.
While these practices increase evaluation cost, they are essential for distinguishing genuine scientific progress from statistical noise.
\end{abstract}

\begin{center}
  \small
  \emoji{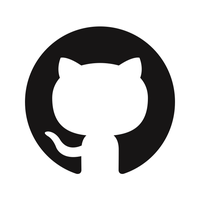}~\href{https://github.com/ASSERT-KTH/agentic-evals-lab}{Inference} \quad
  \emoji{plots/logo_github.png}~\href{https://github.com/ASSERT-KTH/vestige}{Analysis} \quad
  \emoji{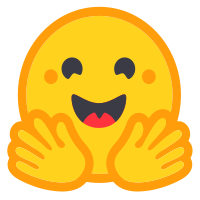}~\href{https://huggingface.co/datasets/ASSERT-KTH/agentic-evals-artifacts}{Data}
\end{center}

\section{Introduction}

Agentic systems that use tools and interact with environments are becoming increasingly capable.
Measuring their performance reliably is essential: evaluation scores guide critical decisions about which models to deploy, whether algorithmic changes provide genuine improvements, and leaderboards are commonly used to quantify how much progress the field is making \citep{kwa2025measuring}.
These decisions have substantial engineering and business consequences.
A reported 3\% improvement might justify adopting a new model, investing in a particular research direction, or making deployment decisions affecting millions of users.
But how reliable are the evaluation scores we use to make these decisions?

Today, most agentic evals follow the approach established for code generation \citep{kulal2019spoc, chen2021evaluating}.
Agents are tested on benchmark tasks like \swebenchverified{} \citep{jimenez2024swebench} and scored using \texttt{pass@1} -- the probability that a task is solved in a single attempt.
Despite the name suggesting a statistical estimator, in practice, most researchers run the agent exactly once per task and report the fraction that succeeded.
This single-run approach has become standard practice across research papers, model releases, and community leaderboards \citep{yang2024sweagent}.

However, single-run evaluation is methodologically unsound for several reasons.
First, estimating \texttt{pass@1} from a single binary outcome per task provides a high-variance estimate of the true success probability.
Second, sampling with temperature $>0$ introduces stochasticity that can produce different outcomes across runs of the same agent on the same task.
Third, beyond sampling, environment interactions might introduce further non-determinism through tool execution or timing effects.

In this paper, we quantify how randomness affects agentic evals.
We conduct ten independent runs (instead of the standard single run) of six agent configurations on \swebenchverified{}, systematically varying models and scaffolds.
In total, we collect \numTrajectories{} agent trajectories, generating over \numTokens{} tokens and \numToolCalls{} tool calls, and systematically analyze their outcomes, performance distributions and divergence points.
We demonstrate substantial randomness in evaluation outcomes: single-run \texttt{pass@1} estimates vary by 2.2 to 6.0 percentage points depending on which run is observed, with standard deviations exceeding 1.5 percentage points even at temperature 0.
This variance persists across all configurations, including theoretically deterministic settings (temperature 0), where non-determinism from inference engines and environments still clearly produces measurable variance.
Through token-level trajectory analysis, we find that runs diverge early, often within the first few percent of tokens, and these initial differences cascade into fundamentally different solution strategies through the autoregressive conditioning mechanism of agentic loops.
Using \texttt{pass@k} (the probability that at least one of $k$ attempts succeeds) \citep{chen2021evaluating} and \texttt{pass\^{}k} (the probability that all $k$ attempts succeed) \citep{yao2025taubench}, we find gaps up to 24.9 percentage points between best-case and worst-case performance, revealing how much success depends on stochastic exploration rather than deterministic problem-solving capability.

These findings have critical implications for interpreting progress in agentic AI.
Many papers claim small improvements based on single-run \texttt{pass@1} scores; our results demonstrate that differences of this magnitude often fall within the natural variance of the evaluation process itself.
A reported one-run improvement from, say, 31\% to 33\% could reflect sampling a favorable run from the same underlying distribution rather than genuine algorithmic progress.
To enable sound evaluation of agentic systems, we recommend three concrete practices: (1) estimate \texttt{pass@1} from multiple independent runs per task, especially when measuring small improvements, (2) using statistical power analysis to determine how many runs to do, and (3) consider multiple metrics like \texttt{pass@k} (optimistic bound) and \texttt{pass\^{}k} (pessimistic bound) with $k>1$ to characterize the full performance envelope.

To summarize, our contributions are:
(1) a large-scale empirical study quantifying variance in agentic evals across three models, two scaffolds, and two temperature settings (\numTrajectories{} trajectories total, \numTokens{} tokens, \numToolCalls{} tool calls);
(2) token-level divergence analysis revealing when and how agent trajectories split into different solution strategies;
(3) characterization of performance bounds using \texttt{pass@k} and \texttt{pass\^{}k} metrics, demonstrating gaps up to 24.9 percentage points between optimistic and pessimistic scenarios; and
(4) concrete, actionable recommendations for reliable evaluation practices that enable sound scientific progress in agentic AI.

\section{Characterizing Randomness in Agentic Evals}

Our goal is to characterize the randomness happening in evals or agentic systems, and understand its sources.
This is essential for interpreting reported scores on leaderboards: total randomness would mean that rankings are unsound and insignificant for decision making.
We design and perform systematic experiments where we run several agents (i.e., model-scaffold pairs) ten times each and analyze the distribution of outcomes and the agentic trajectories.
We perform these experiments with both theoretically deterministic sampling (temperature=0.0), as well as with the sampling hyper-parameters suggested by the authors of each model, which is the temperature typically used in leaderboards.

\subsection{Experimental Setup}

We consider agentic coding as the domain of choice for our experiments, as it is one of the most popular and active domains for agentic research.
Agentic coding tasks are often highlighted in model cards and used to make claims about trends in AI development \citep{kwa2025measuring}.
Particularly, we focus on the software engineering issue resolution tasks from the \swebenchverified{} benchmark \citep{jimenez2024swebench}.
This is the most widely used benchmark for agentic coding, and is massively used in model cards and research papers.
In \swebenchverified{}, the agents are tasked with resolving a GitHub issue and their success is validated through automated unit tests.

We exhaustively evaluate six different agents, where an agent is defined as a model-scaffold pairs. We consider the following models:
\begin{itemize}
  \item \texttt{Qwen/Qwen3-32B} \cite{yang2025qwen3} is a medium sized model commonly used by researchers in agentic coding experiments with, at the time of writing, over 1.2M downloads over 300 fine-tuned versions available on Hugging Face. This model is a common model used for research on agentic models \citep{deepswe2025, tang2025beyond, cao2025skyrl, qian2025userrl}.
  \item \texttt{agentica-org/DeepSWE-preview} \cite{deepswe2025} is a fine-tuned variant of \texttt{Qwen/Qwen3-32B} specifically for agentic coding. For us, this model is meant to represent fine-tuned  models for agentic coding.
  \item \texttt{mistralai/Devstral-2-123B-Instruct-2512} \citep{rastogi2025devstral} is a large open-weights model specifically fine-tuned for agentic coding, achieving state-of-the-art performance amongst open-weights models on \swebenchverified{}.
  It enables us to give perspective on the \texttt{DeepSWE-preview} results, both models being fine-tuned for agentic coding.
\end{itemize}

And the following scaffolds:
\begin{itemize}
  \item \texttt{nano-agent} is our own minimal scaffold for agentic coding experiments, providing a simple yet functional environment for agent-task interaction. We use it because it is guaranteed to not have been used in the training of any models we evaluate, allowing us to reason about scaffold independence with guarantees.
  \item \texttt{R2E-Gym} is a code agent scaffold proposed by \citeauthor{jain2025r2e} and used during the training of \texttt{DeepSWE-preview}. This scaffold is more feature-rich than \texttt{nano-agent} and allows us to inspect the effect of evaluating a model on its specific training scaffold.
\end{itemize}

Our selection of models and scaffolds is designed to mitigate potential implementation-specific artifacts that could confound our variance measurements.
Specifically, we introduce diversity across two dimensions:
(1) scaffold implementation, using both our own \texttt{nano-agent} and the independently developed \texttt{R2E-Gym}, and
(2) model deployment infrastructure, where \texttt{Qwen3-32B} and \texttt{DeepSWE-preview} are deployed locally with vLLM while \texttt{Devstral-2} is accessed through Mistral's hosted API.
This orthogonal variation ensures that observed variance patterns are not artifacts of bugs or idiosyncrasies in any single implementation, but generalize well.

Both scaffolds use append-only conversation contexts without any context truncation, compaction, or summarization strategies.
This property is important for our trajectory divergence analysis and token accounting methodology (see \Cref{sec:eval_metrics}), as these measurements assume the complete conversation history is preserved throughout the interaction. Studying the randomness induced by compaction is left to future work.

\subsection{Agent Trajectories}
\label{sec:trajectories}

To understand the mechanisms underlying randomness in evals, we need to analyze agent trajectories at the token level.
We formalize the concepts of trajectories and token usage below.
These definitions apply to scaffolds with append-only conversation contexts.

An agentic run consists of $K$ interaction steps. At each step $k$, the model receives a context $C_k$ containing all prior messages and generates a response $G_k$ (which may include reasoning tokens, text, and tool calls). The environment then provides a response $R_k$ (e.g., tool execution results).

\textbf{Trajectory.} We define the trajectory $\tau_j$ for run $j$ as the complete linearized sequence of all messages in chronological order, including both model-generated tokens and environment-generated tokens  (tool responses):
\begin{equation}
\tau_j = C_1 \oplus G_1 \oplus R_1 \oplus G_2 \oplus R_2 \oplus \cdots \oplus G_K \oplus R_K
\end{equation}
where $\oplus$ denotes concatenation and $C_1$ includes the initial system and user prompt.
The trajectory includes both model-generated tokens and environment-generated tokens, as both influence subsequent model behavior through autoregressive conditioning.

\subsection{Metrics}
\label{sec:eval_metrics}

We employ several complementary metrics to characterize agent performance, each capturing different aspects of agent behavior.
We consider a benchmark with $N$ tasks and $m$ independent evaluation runs on each task (in our case, $N=500$ and $m=10$). Let $c_i$ denote the number of successful attempts for task $i$ across all $m$ runs.

\textbf{Single-run resolution rate:}
Let $r_j$ denote the resolution rate from run $j$, computed as the fraction of tasks solved in that run:
\begin{equation}
  r_j = \frac{|\{i : \text{task } i \text{ solved in run } j\}|}{N}
\end{equation}
When we perform $m$ independent runs, we obtain $m$ different values $r_1, r_2, \ldots, r_m$.
The mean $\overline{r} = \frac{1}{m}\sum_{j=1}^{m} r_j$ and standard deviation of these values quantify the expected performance and run-to-run variability.
In \Cref{tab:swebench_scores}, we report these statistics to characterize the distribution of outcomes across runs.

\textbf{\texttt{pass@k} and \texttt{pass\^{}k}}:
With multiple evaluation runs on each task, we can compute two complementary metrics to characterize agent capabilities.
The \texttt{pass@k} metric \citep{chen2021evaluating} estimates the probability that at least one of $k$ randomly selected attempts succeeds.
The \texttt{pass\^{}k} metric \citep{yao2025taubench} estimates the probability that all $k$ attempts succeed:
\begin{equation}
\text{pass}@k = \frac{1}{N}\sum_{i=1}^{N}\left[1 - \frac{\binom{m-c_i}{k}}{\binom{m}{k}}\right]
\quad\text{and}\quad
\text{pass}^{\wedge}k = \frac{1}{N}\sum_{i=1}^{N}\left[\frac{\binom{c_i}{k}}{\binom{m}{k}}\right]
\end{equation}
where $\binom{a}{b}$ denotes the binomial coefficient.

The \texttt{pass@k} metric answers: ``If we randomly select $k$ of our $m$ attempts for each task, what fraction of tasks would be solved at least once?''
For $k=1$, this estimator reduces to $\frac{1}{N}\sum_{i=1}^{N} \frac{c_i}{m} = \overline{r}$, which is exactly the mean of single-run resolution rates.
Thus, $\texttt{pass@1} = \overline{r}$ represents the pooled estimate of first-attempt success probability across multiple runs.
For $k>1$, the \texttt{pass@k} estimator properly accounts for the combinatorics of sampling and differs from simply averaging empirical success rates.
It also represents an optimistic bound of model capabilities.
Complementarily, \texttt{pass\^{}k} measures consistency and robustness, also representing a pessimistic bound of model capabilities.
A high \texttt{pass\^{}k} indicates that the agent reliably solves tasks across multiple attempts, while a low \texttt{pass\^{}k} relative to \texttt{pass@k} suggests success depends heavily on stochastic exploration.

In many related works:
1) $\texttt{pass@1}$ is reported as the only metric;
2) too often based on a single run;
3) too rarely, the number of runs used to estimate it is reported.

\textbf{First token divergence ($\tau_{\text{div}}$):}
To understand when and how agent runs diverge, we measure the first position at which two trajectories differ.
Using the trajectory formalism from \Cref{sec:trajectories}, for two runs $i$ and $j$ on the same task with tokenized trajectories $\tau_i = [t_1^i, t_2^i, \ldots]$ and $\tau_j = [t_1^j, t_2^j, \ldots]$, we define:
\begin{equation}
\tau_{\text{div}}(i,j) = \min\{k : t_k^i \neq t_k^j\}
\end{equation}
This metric captures when trajectories begin to explore different solution paths, which is critical for understanding variance propagation in agentic evals.

\subsection{Experimental Results}

Our experiment yields \numTrajectories{} agent trajectories in total, from 120 experimental runs (6 configurations $\times$ 10 runs each $\times$ 2 scaffolds).
These runs consumed \numTokens{} tokens, generating \numToolCalls{} tool calls.
In this section, we quantify the randomness in evaluation outcomes, as well as try to understand the mechanisms behind it.

\subsubsection{Quantifying Randomness in Evaluation Outcomes}
\label{sec:rq1}
\begin{table}[t]
  \caption{Resolution rates across 10 independent evals on \swebenchverified{}. Each row shows statistics of $r_j$ (single-run resolution rates) computed over 10 separate runs with identical configuration. Mean values ($\overline{r}$) are equivalent to \texttt{pass@1} estimated by pooling all runs (see \Cref{sec:eval_metrics}), while standard deviation quantifies run-to-run variability. The substantial ranges (min to max) demonstrate the presence of randomness in evaluation outcomes, even with identical settings and temperature 0.}
  \label{tab:swebench_scores}
  \centering
  \small
  \setlength{\tabcolsep}{4pt}
  \begin{tabular}{@{}l c ccc ccc@{}}
  \toprule
  \multirow{2}{*}{Model} & \multirow{2}{*}{Temp} & \multicolumn{3}{c}{\texttt{nano-agent}} & \multicolumn{3}{c}{\texttt{r2e-gym}} \\
  \cmidrule(lr){3-5} \cmidrule(lr){6-8}
   &  & $\overline{r}\pm$Std & Min & Max & $\overline{r}\pm$Std & Min & Max \\
  \midrule
  \texttt{Qwen3-32B} & 0.6 & $16.4\% \pm 0.7\%$ & 15.0\% & 17.2\% & $23.9 \% \pm 1.4\%$ & 21.4\% & 26.4\% \\
  \texttt{DeepSWE-preview} & 1.0 & $31.4 \% \pm 1.0\%$ & 28.8\% & 32.4\% & $34.4 \% \pm 1.5\%$ & 31.6\% & 37.0\% \\
  \texttt{Devstral-2} & 0.2 & $63.5 \% \pm 1.1\%$ & 61.8\% & 65.0\% & $34.9 \% \pm 1.5\%$ & 32.2\% & 37.0\% \\
  \texttt{Qwen3-32B} & 0.0 & $16.4 \% \pm 1.2\%$ & 14.4\% & 18.0\% & $22.3 \% \pm 1.8\%$ & 19.8\% & 25.2\% \\
  \texttt{DeepSWE-preview} & 0.0 & $20.4 \% \pm 1.0\%$ & 18.2\% & 21.4\% & $19.2 \% \pm 1.5\%$ & 17.0\% & 21.6\% \\
  \texttt{Devstral-2} & 0.0 & $63.8 \% \pm 1.6\%$ & 60.6\% & 66.6\% & $35.4 \% \pm 1.7\%$ & 32.0\% & 37.8\% \\
  \bottomrule
  \end{tabular}
  \end{table}
  
\Cref{tab:swebench_scores} presents the single-run resolution rates $r_j$ (percentage of tasks successfully resolved) for each model-scaffold-temperature combination, aggregated across 10 independent evals per agent under test.
We report the mean ($\overline{r} = \texttt{pass@1}$), standard deviation, minimum, and maximum values to characterize the distribution of outcomes.
Across all conditions, we observe substantial run-to-run variability.
For example, \texttt{DeepSWE-preview} on \texttt{nano-agent} with temperature 1.0 achieves $\overline{r} = 31.4 \pm 1.0\%$ (\texttt{pass@1}), with individual runs ranging from 28.8\% to 32.4\% (a 3.6 percentage point spread.
Similarly, \texttt{Qwen3-32B} on \texttt{R2E-Gym} with temperature 0.6 shows a mean of $23.9 \pm 1.4\%$, ranging from 21.4\% to 26.4\% (a 5.0 percentage point spread).
Across all twelve configurations in the table, the ranges span 2.2 to 6.0 percentage points, representing substantial variability where a single run could report performance anywhere within this window.
This variability is significant enough that improvements measured with a single run might be purely due to randomness in the evaluation process rather than a genuine improvement.

\textbf{``Deterministic'' sampling.}
In theory, evals can be run deterministically with temperature 0. In practice, determinism is not achievable \citep{yuan2025understanding, he2025nondeterminism}: modern LLM inference engines introduce various sources of non-determinism including floating-point precision, parallelization, hardware-specific optimizations, and batching strategies.
We study the extent of the problem at temperature 0 (bottom half of the table).
Clearly, this variance persists even with theoretically deterministic sampling (temperature 0.0).
For instance, \texttt{DeepSWE-preview} on \texttt{nano-agent} (temp zero) achieves $20.4 \pm 1.0\%$ (range: 18.2\%--21.4\%), and \texttt{Qwen3-32B} on \texttt{R2E-Gym} achieves $22.3 \pm 1.8\%$ (range: 19.8\%--25.2\%).
Counter-intuitively, the variance never decreases and sometimes increases with temperature zero (eg 0.7\% variance for Qwen3-32B x nano-agent at temperature 0.6 to 1.2\% variance at temperature 0).
To sum up, temperature 0 does not result in determinism.

\textbf{Statistical significance of temperature effects.}
The impact of temperature on performance varies significantly across models, highlighting the importance of statistical testing rather than relying solely on single-runs or mean differences.
For \texttt{DeepSWE-preview}, temperature 1.0 achieves 31.4\% $\pm$ 1.0\% on \texttt{nano-agent} versus 20.4\% $\pm$ 1.0\% at temperature 0.0 (11.0 percentage point difference).
Given the standard deviations, this difference is statistically significant, indicating that stochastic exploration genuinely improves the problem-solving capability of this model.
In contrast, \texttt{Devstral-2} shows no statistically significant difference: on \texttt{nano-agent}, temperature 0.2 achieves 63.5\% $\pm$ 1.1\% versus 63.8\% $\pm$ 1.6\% at temperature 0.0.
Despite the slightly higher mean at temperature 0.0, the 0.3 percentage point difference is well within the noise given the large standard deviations (1.1\% and 1.6\%).
These examples demonstrate that statistical testing is essential to distinguish genuine effects from random variation, as we further develop in \Cref{sec:recommendations} and \Cref{app:power-analysis}.

\textbf{Implications.} Our results demonstrate substantial randomness in agentic evaluation outcomes, even under identical configurations.
The practical implications are significant: single-run scores in technical reports and articles can be misleading without statistical bounds.
Consider evaluating whether a model improves performance. Observing a 3 percentage point increase on a single run could be purely due to randomness in the evaluation process rather than a genuine improvement.
This affects how we interpret progress in the field: a newly released model claiming a 2-3 point improvement over its predecessor might simply be reporting a favorable outcome from the same underlying distribution.
To enable sound decision-making, for research, development, or deployment, evaluation results should be reported with variance estimates over multiple runs rather than point estimates from single runs.

\subsubsection{What do \texttt{pass@1}, \texttt{pass@k}, and \texttt{pass\^{}k} reveal about agent randomness?}

\begin{figure}[t]
\centering
\begin{subfigure}[b]{0.48\textwidth}
\centering
\includegraphics[width=\textwidth]{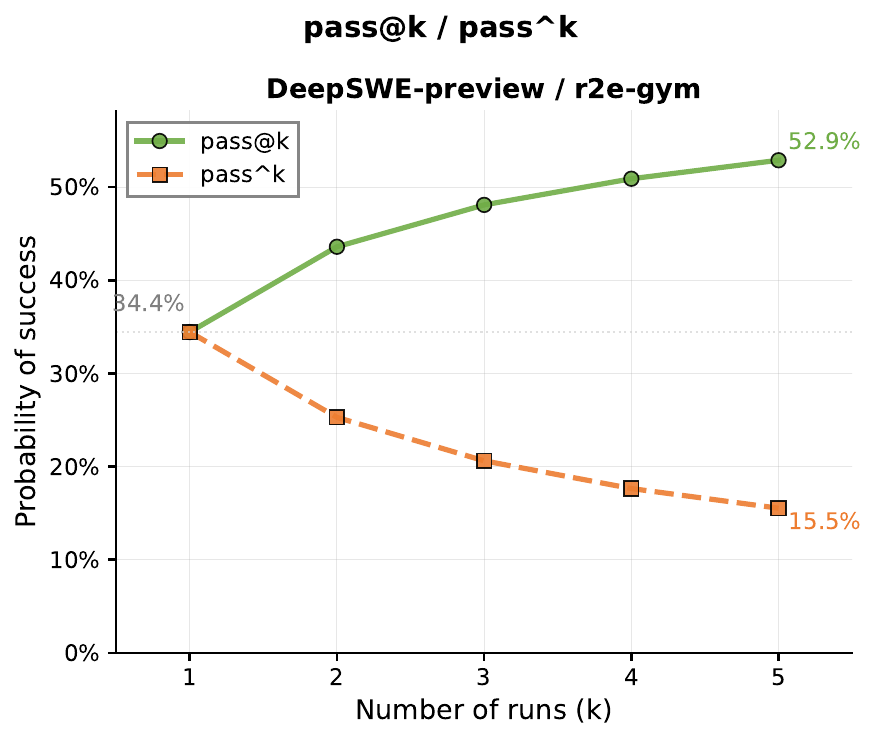}
\end{subfigure}
\hfill
\begin{subfigure}[b]{0.48\textwidth}
\centering
\includegraphics[width=\textwidth]{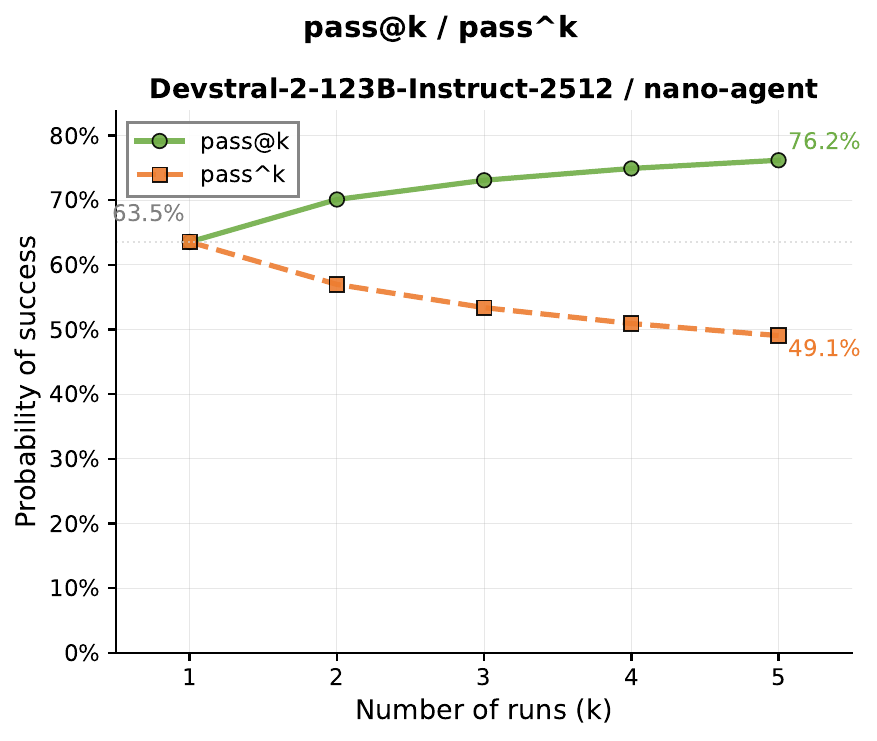}
\end{subfigure}
\caption{Performance bounds revealed by \texttt{pass@k} and \texttt{pass\^{}k} for \texttt{DeepSWE-preview} on \texttt{r2e-gym} and \texttt{Devstral-2} on \texttt{nano-agent}. The vertical distance between curves quantifies how much performance depends on random choices. \texttt{DeepSWE-preview} exhibits wider gaps (high sensitivity to randomness), while \texttt{Devstral-2} shows narrower gaps (more consistent solutions), though both demonstrate substantial dependence on stochastic exploration as $k$ increases.}
\label{fig:pass_at_k}
\end{figure}

Randomness in agentic trajectories has a good aspect: it can be leveraged to increase performance via retrying.
We now examine the extremes.
What is the best performance we could achieve if we exploit this randomness via retries? 
What is the worst performance we can expect when randomness is maximally unfavorable?

We analyze three metrics that capture different aspects of this question, see \Cref{sec:eval_metrics}.
A large (\texttt{pass@k} - \texttt{pass@1}) gap indicates the agent has the potential to solve many tasks, but needs multiple attempts to find the right path.
A large (\texttt{pass@1} - \texttt{pass\^{}k}) gap indicates the agent's success is highly sensitive to which random choices are made and might not be able to reliably produce consistent solutions.
Together, these metrics bound the agent's capabilities.

Figure \ref{fig:pass_at_k} shows those three metrics for two modelxscafoold pairs.
It reveals substantial gaps between these bounds, exposing how much agent performance depends on randomness.
For \texttt{DeepSWE-preview} on \texttt{r2e-gym}, the first-attempt success probability (\texttt{pass@1}) is 34.4\%, but with five retries, performance reaches 52.9\% (\texttt{pass@5}), an 18.5 percentage point improvement representing the optimistic potential.
At the other extreme, only 15.5\% of tasks (\texttt{pass\^{}5}) are solved consistently across five attempts, which is less than half of the \texttt{pass@1} rate.
This 18.9 percentage point gap between \texttt{pass@1} (34.4\%) and \texttt{pass\^{}5} (15.5\%) reveals that part of the agent's capability depends on favorable random choices.

This pattern generalizes across configurations, though the magnitude varies.
Consider \Cref{fig:pass_at_k}, where we show the \texttt{pass@k} and \texttt{pass\^{}k} curves for \texttt{DeepSWE-preview} on \texttt{r2e-gym} and \texttt{Devstral-2} on \texttt{nano-agent} (other model-scaffold pairs are provided in the appendix)).
\texttt{Devstral-2} on \texttt{nano-agent} shows a narrower range: \texttt{pass@1} is 63.5\%, \texttt{pass@5} reaches 76.2\% (12.7 point gap to optimistic bound), and \texttt{pass\^{}5} is 49.1\% (14.4 point gap to pessimistic bound).
These narrower gaps indicate that higher-performing models exhibit more consistent solution strategies, yet they still benefit significantly from stochastic exploration.
Across all twelve configurations, the maximum improvement from \texttt{pass@1} to \texttt{pass@5} is 24.9 percentage points (\texttt{Devstral-2} on \texttt{r2e-gym} at temperature 0), demonstrating that scaffold choice and model-scaffold interaction significantly impact the degree of stochastic dependence.

In summary, the gap between \texttt{pass@k} (optimistic bound) and \texttt{pass\^{}k} (pessimistic bound) quantifies how much agent performance depends on favorable stochastic exploration.
This dependence is substantial across all configurations and demonstrate that randomness is not a minor perturbation but a fundamental component of agent performance, with implications for evaluation methodologies and interpretation of results.

\subsubsection{Understanding the Mechanism: When Do Trajectories Diverge?}

\begin{figure}[t]
  \centering
  \begin{subfigure}[b]{0.32\textwidth}
  \centering
  \includegraphics[width=\textwidth]{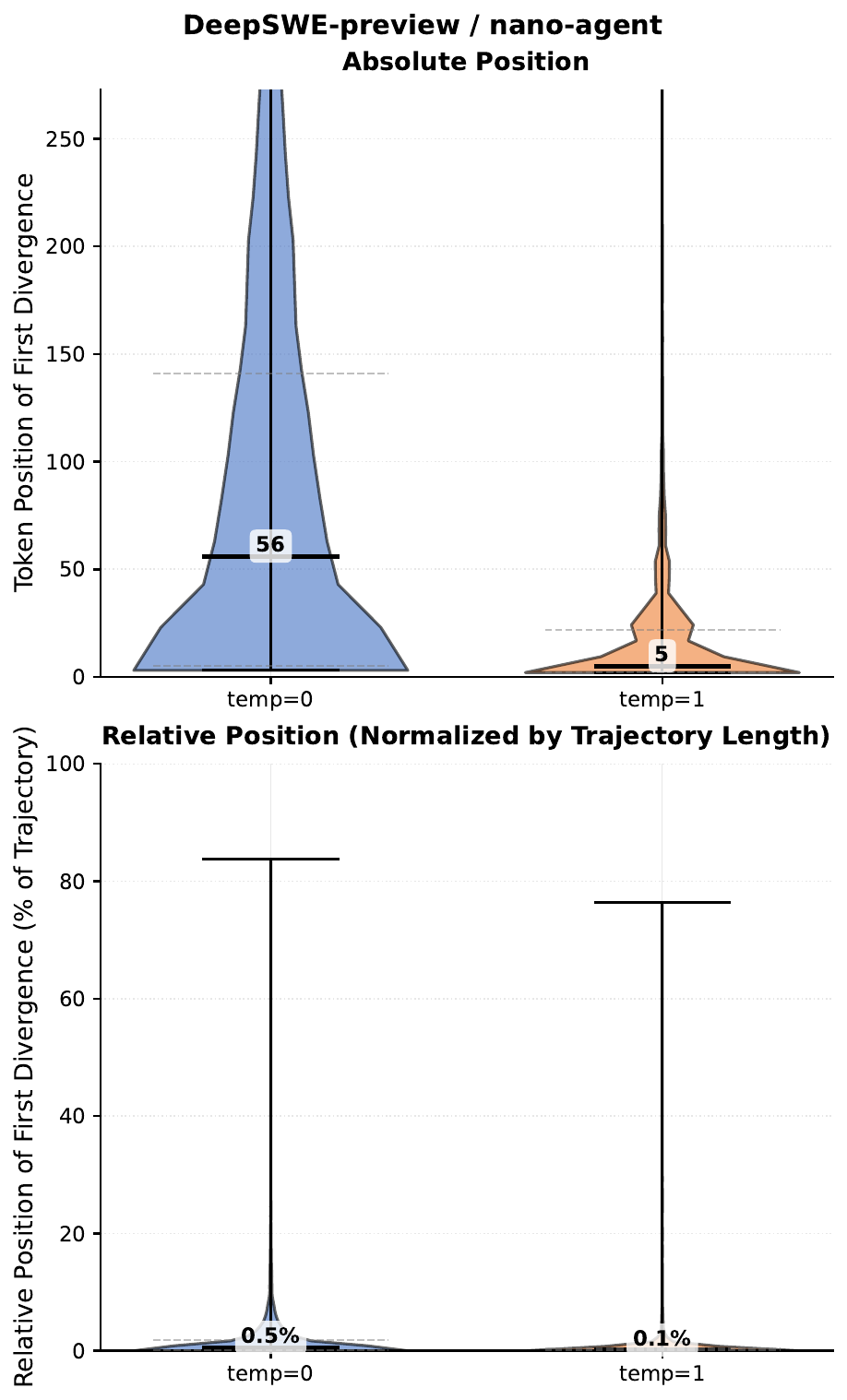}
  \end{subfigure}
  \hfill
  \begin{subfigure}[b]{0.32\textwidth}
  \centering
  \includegraphics[width=\textwidth]{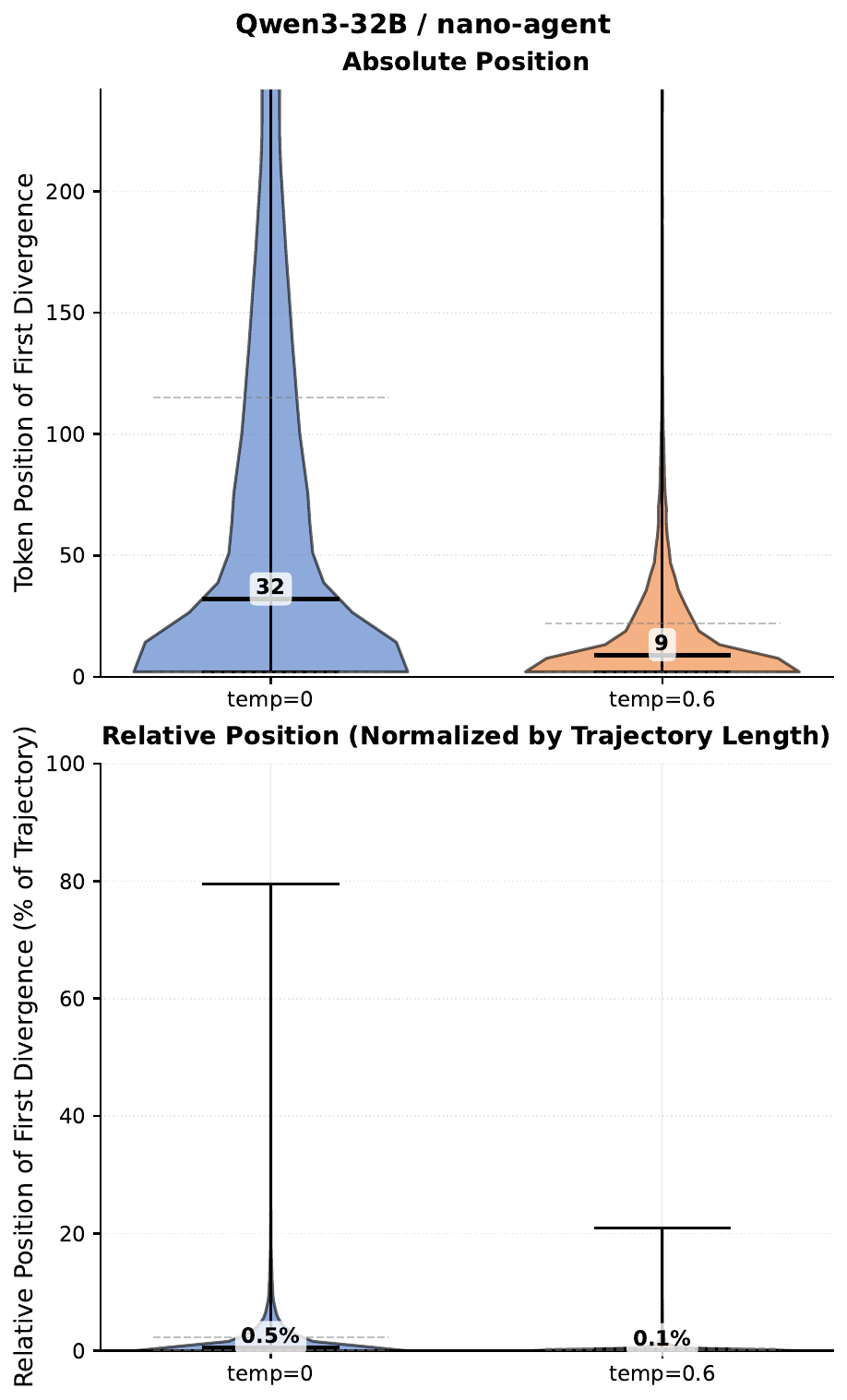}
  \end{subfigure}
  \begin{subfigure}[b]{0.32\textwidth}
  \centering
  \includegraphics[width=\textwidth]{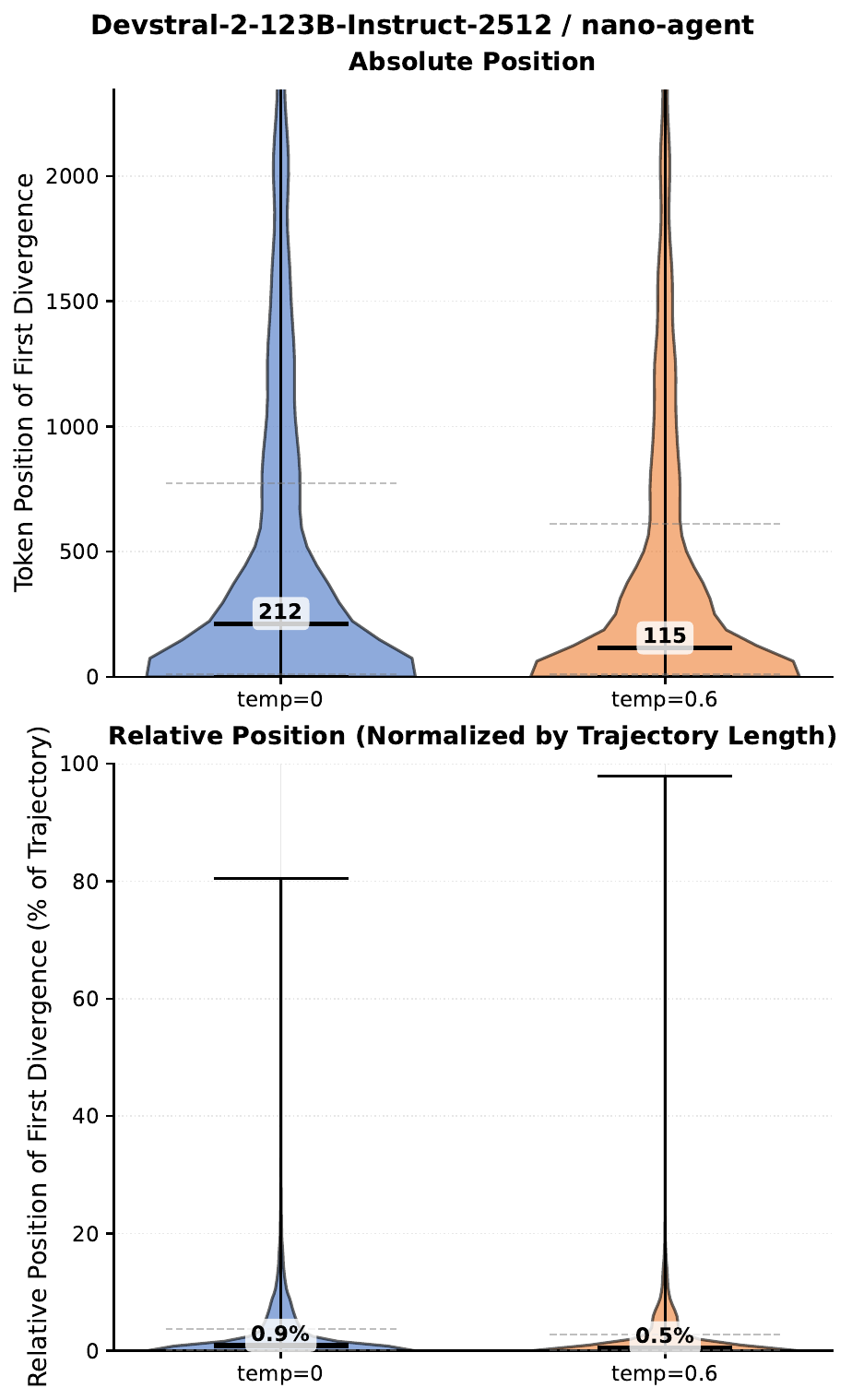}
  \end{subfigure}
  \caption{Distribution of first token divergence across different models under \texttt{nano-agent}. In blue, we show the distributions with temperature 0, while in orange we show the distributions with the suggested temperatures. On top, we plot by absolute token position, while on bottom, we plot by relative position (percentage through the trajectory). The distributions are shown for all pairs of divergent runs, one per model-scaffold pair.}
  \label{fig:variance_comparison}
\end{figure}

To qualitatively understand the underlying randomness in agentic outcomes, we analyze when and how runs diverge at the trajectory level.
Once a single token differs between two runs, the probability distribution (logits) computed by the LLM for subsequent tokens also changes, since the model conditions on the full context including the divergent token.
This creates a butterfly effect: a single early difference can propagate through the trajectory, affecting ever more tokens, tool calls, observations, and ultimately the final outcome.

\Cref{fig:variance_comparison} shows the distribution of first token divergence position with (blue) and without (red) deterministic sampling, across all pairs of runs, one per model-scaffold pair.
The top plots show the distribution of divergence by absolute token position, while the bottom plot shows the distribution by relative position (percentage through the trajectory), revealing whether divergence is consistently early regardless of trajectory length.

\textbf{Early divergence.} \Cref{fig:variance_comparison}  shows that the distribution of first token divergence reveals that trajectories typically diverge very early, within the first tokens (top figure) immediately after the common system and user prompts.
Relative to the size of the trace, the first divergence always happen in the 1\% of the total trajectory (bottom of the figure).
For example, for \texttt{DeepSWE-preview} on \texttt{nano-agent}, the median first token divergence occurs at token position 5 with default temperature (1.0), that is at 0.5\% of the total trajectory length.

\textbf{Temperature effect on trajectories.} Deterministic sampling (temperature 0.0) shifts divergence substantially later, as expected.
For \texttt{DeepSWE-preview} on \texttt{nano-agent}, the median first token divergence position increases from 5 (default temperature 1.0) to 56 when using temperature 0.0.
Similarly, \texttt{Qwen3-32B} on \texttt{nano-agent} exhibits median divergence at token 9 with default temperature (0.6), increasing to token 32 at temperature 0.0.
The same phenomenon is observed for \texttt{Devstral-2}.
So, temperature does have a little positive impact.

Yet, confirming the results of \Cref{sec:rq1}, deterministic sampling only delays divergence, but does not suppress it.
Since trajectory divergence happens early, we expect that the longer the trajectories, the more likely they are to also diverge semantically, because of cascading effects in the next token probability distributions.
As the community moves towards more complex and long-horizon tasks, the importance of measuring variance will only increase.

\textbf{Case study.} \Cref{fig:divergence_example} illustrates a concrete example of how early divergence in reasoning cascades into drastically different outcomes.
On task \texttt{django\_\_django-9296}, two independent runs of \texttt{Qwen3-32B} on \texttt{nano-agent} at temperature 0 generated identical reasoning for the first 93 tokens.
At token 94, within the model's internal reasoning trace, a subtle difference emerged: run 1 reasoned ``Let me search for the Paginator class'' while run 2 reasoned ``Let me check the Django source code.''
This seemingly minor phrasing difference propagated through the autoregressive conditioning to lead to different first tool calls, where the first searches for the \texttt{Paginator} class in a specific file while the other searches in an entire directory, leading to different tool call outputs.
At the end, this cascades into fundamentally different problem-solving strategies, and with opposite outcomes.
Run 1 found the correct file but applied an incorrect patch that inserted the new method in the wrong location breaking Python syntax, failing the task.
Run 2's broader search led to more careful analysis of the code structure, ultimately finding the correct insertion point and successfully resolving the task.
Even at temperature 0, randomness causes divergent reasoning that compounds through autoregressive conditioning into fundamentally different problem-solving strategies with opposite outcomes.

\begin{figure}[t]
\centering
\small

\begin{tikzpicture}
\node[fill=gray!15, rounded corners, inner sep=4pt, font=\small] {
\textbf{Task:} \texttt{django\_\_django-9296} \quad
\textbf{Model:} \texttt{Qwen3-32B} (temp=0) \quad
\textbf{Scaffold:} \texttt{nano-agent}
};
\end{tikzpicture}

\vspace{0.3em}

\fcolorbox{black!20}{yellow!5}{\begin{minipage}{0.97\textwidth}

{\ttfamily\scriptsize
<|im\_start|>assistant\\
<think>\\
Okay, let's see. The user wants to modify the Paginator class to implement the \_\_iter\_\_ method so that iterating over it goes through all pages using page\_range. The current approach uses page\_range, but they think it's more natural to use Python's standard iteration.\\[0.3em]
First, I need to find where the Paginator class is defined. Since the user mentioned it's related to Django, the Paginator is likely in Django's core.
}

\vspace{0.2em}
\begin{tikzpicture}
\node[anchor=west, font=\scriptsize, text width=13.5cm] (run1a) at (0,0) {
\textbf{Run 1:}~~~\colorbox{red!15}{\strut Let me search for the Paginator class in the Django source code.}
};
\node[anchor=west, font=\scriptsize, text=black!60, text width=13cm] (run1b) at (0.45,-0.4) {
\textit{...I'll use the shell command to search for 'class Paginator' in the Django repository...}
};
\node[anchor=west, font=\scriptsize, text width=13.5cm] (run2a) at (0,-0.9) {
\textbf{Run 2:}~~~\colorbox{green!15}{\strut Let me check the Django source code. Using the shell tool, I can search...}
};
\node[anchor=west, font=\scriptsize, text=black!60, text width=13cm] (run2b) at (0.45,-1.3) {
\textit{...I'll start by using the 'find' command to locate the file. Alternatively, using 'rg -l'...}
};
\end{tikzpicture}

{\ttfamily\scriptsize
</think>
}

\end{minipage}}

\vspace{0.3em}

\fcolorbox{black!20}{blue!5}{\begin{minipage}{0.97\textwidth}

\begin{tikzpicture}
\node[anchor=west, font=\scriptsize] (run1label) at (0,0.55) {
\textbf{Run 1:}
};
\node[anchor=west, font=\ttfamily\scriptsize] (run1j1) at (1.0,0.55) {\{};
\node[anchor=west, font=\ttfamily\scriptsize] (run1j2) at (1.2,0.2) {"name": "shell",};
\node[anchor=west, font=\ttfamily\scriptsize] (run1j3) at (1.2,-0.1) {"arguments": \{"cmd": "rg ... \colorbox{red!25}{\strut django/core/paginator.py}"\}};
\node[anchor=west, font=\ttfamily\scriptsize] (run1j5) at (1.0,-0.4) {\}};
\node[anchor=west, font=\scriptsize, text=black!70, text width=10.5cm] (out1a) at (1.0, -0.7) {
\textit{Approach:} Assumes file path, searches within specific file
};
\node[anchor=west, font=\scriptsize, text=red!70, text width=10.5cm] (out1b) at (1.0, -1.0) {
\textit{Outcome:} Applies patch at wrong location $\rightarrow$ \textbf{Fails}
};

\draw[gray!40, dashed] (-0.1, -1.25) -- (13, -1.25);

\node[anchor=west, font=\scriptsize] (run2label) at (0,-1.5) {
\textbf{Run 2:}
};
\node[anchor=west, font=\ttfamily\scriptsize] (run2j1) at (1.0,-1.5) {\{};
\node[anchor=west, font=\ttfamily\scriptsize] (run2j2) at (1.2,-1.85) {"name": "shell",};
\node[anchor=west, font=\ttfamily\scriptsize] (run2j3) at (1.2,-2.15) {"arguments": \{"cmd": "rg ... \colorbox{green!25}{\strut django/core/}"\}};
\node[anchor=west, font=\ttfamily\scriptsize] (run2j5) at (1.0,-2.45) {\}};
\node[anchor=west, font=\scriptsize, text=black!70, text width=10.5cm] (out2a) at (1.0, -2.75) {
\textit{Approach:} Searches directory, explores code more widely
};
\node[anchor=west, font=\scriptsize, text=green!60!black, text width=10.5cm] (out2b) at (1.0, -3.05) {
\textit{Outcome:} Finds correct buggy location $\rightarrow$ \textbf{Succeeds}
};
\end{tikzpicture}

\end{minipage}}

\caption{A subtle reasoning divergence at token 94 cascades into opposite outcomes. Both runs share identical reasoning through the first paragraph, understanding the task of adding an \texttt{\_\_iter\_\_} method to Django's \texttt{Paginator} class. At token 94, the reasoning diverges: run 1 reasons ``Let me search...'' while run 2 reasons ``Let me check...Using the shell tool...''. This difference leads to a different first tool call, which propagates through subsequent steps, with only run 2 succeeding. Even at temperature 0, non-determinism causes trajectory divergence that compounds into fundamentally different problem-solving strategies.}
\label{fig:divergence_example}
\end{figure}

To sum up, early divergences have important implications for long-horizon agentic tasks.
Since divergence occurs early and propagates through the remainder of the trajectory via the autoregressive conditioning mechanism, longer trajectories exhibit amplified variance.  In agentic evals, more than zero-shot prompting, small initial perturbations lead to increasingly divergent outcomes as the trajectory lengthens.

\section{Implications and Mitigation Strategies}

\subsection{False sense of progress}
The variance documented in this paper has immediate consequences for how we interpret progress in agentic systems.
Single-run evaluations can lead to researchers not being able to determine whether observed differences represent genuine capability gaps or merely different samples from overlapping performance distributions.
This problem extends beyond individual papers to affect the broader scientific ecosystem.
Leaderboards that rank systems based on single-run scores may reflect evaluation noise rather than true capability ordering.
Research directions may be chosen based on apparent improvements that are not statistically distinguishable from noise.
Organizations making deployment decisions, deciding whether to adopt a new model or agentic tool, or allocating engineering resources, face similar challenges.
The scores guiding these decisions may not reliably reflect underlying performance differences.

The problem is particularly acute because evaluation practices have not kept pace with the evolution of agentic systems.
While the \texttt{pass@1} metric originated in code generation settings with relatively short, independent generations like HumanEval \citep{chen2021evaluating}, agentic tasks involve long-horizon, multi-step trajectories where early divergence cascades through subsequent actions.
Our trajectory analysis (\Cref{sec:trajectories}) demonstrates that this cascading effect might amplify variance.
As the field moves toward longer-horizon tasks with more complex tool use, this amplification effect is likely to intensify further.

\subsection{Recommendations}
\label{sec:recommendations}

\begin{table}[t]
  \centering
  \small
  \caption{Required runs per agent to detect improvements at different variance and significance levels (power = 80\%).}
  \label{tab:power-analysis-requirements}
  \begin{tabular}{@{}ccrrr@{}}
  \toprule
  \textbf{Improvement} & \textbf{Std Dev ($\sigma$)} & \textbf{$n$ ($p<0.05$)} & \textbf{$n$ ($p<0.01$)} & \textbf{$n$ ($p<0.001$)} \\
  \midrule
  1\% & 0.7\% &   8 &  12 &  17 \\
  1\% & 1.5\% &  36 &  53 &  77 \\
  1\% & 1.8\% &  51 &  76 & 111 \\
  \midrule
  2\% & 0.7\% &   2 &   3 &   5 \\
  2\% & 1.5\% &   9 &  14 &  20 \\
  2\% & 1.8\% &  13 &  19 &  28 \\
  \midrule
  5\% & 0.7\% &   1 &   1 &   1 \\
  5\% & 1.5\% &   2 &   3 &   4 \\
  5\% & 1.8\% &   3 &   4 &   5 \\
  \midrule
  10\% & 0.7\% &   1 &   1 &   1 \\
  10\% & 1.5\% &   1 &   1 &   1 \\
  10\% & 1.8\% &   1 &   1 &   2 \\
  \bottomrule
  \end{tabular}
\end{table}

To enable reliable evaluation, we recommend running multiple runs per agent under test, and estimating the performance metrics from them.
The required number of runs depends on the magnitude of improvement to detect and the desired statistical power (the probability of correctly identifying a real improvement when it exists).
\Cref{tab:power-analysis-requirements} shows the required number of runs per agent under test for detecting different improvement magnitudes at various significance levels, assuming a normal distribution of randomness.\footnote{We verified this assumption via Shapiro-Wilk tests \citep{shapiro1965analysis} across all 12 configuration--scaffold groups ($n=10$ runs each); 11 of 12 passed at $p \geq 0.05$. See \Cref{app:normality} for details.} The table presents three variance scenarios corresponding to the minimum ($\sigma = 0.7\%$), median ($\sigma = 1.5\%$), and maximum ($\sigma = 1.8\%$) standard deviations observed across our experiments.

Detecting a 2\% improvement at $p < 0.05$ with 80\% power requires approximately 9 runs per agent under test, while detecting a 1\% improvement requires 36 runs.
A study like ours, in which 10 runs are made per agent under test, can reliably detect improvements $\geq 2$ percentage points but not smaller effects.

Detecting a 1\% improvement at median variance levels requires 36 runs, while the same detection at the lowest observed variance ($\sigma = 0.7\%$) would require only 8 runs.
On the other hand, detecting large improvements (e.g., 10\%) can be done with a much smaller number of runs and, depending on the desired statistical power and significance threshold, might even be possible with single-runs.
Further analysis can be found in \Cref{app:power-analysis}.

We also suggest characterizing the performance envelope: by always reporting \texttt{pass@1} (expected performance), \texttt{pass@k} (optimistic bound with retries), and \texttt{pass\^{}k} (pessimistic consistency bound).
The \texttt{pass@k}-\texttt{pass\^{}k} gaps reveal how much stochasticity might be detrimental or beneficial for the agent under test.

We acknowledge that multiple runs increase cost, which is a valid concern for GPU-poor organizations, in particular in academic settings like ours.
However, this investment is necessary to avoid long term costs due to poorly informed decisions, at local and systemic levels.

\section{Related Work}

\subsection{Randomness in Large Language Models}

Recent work has identified multiple sources of non-determinism in large language models.
At the infrastructure level, \citet{yuan2025understanding} and \citet{he2025nondeterminism} demonstrate that non-associative floating point operations, rounding errors, hardware configuration, and batch size variations impact reproducibility at temperature 0.
For code generation specifically, \citet{ouyang2025empirical} find that repeated queries yield different implementations even with greedy sampling.
Prompt sensitivity represents another major source of variance. \citet{zhuo2024prosa, sclar2024quantifying, andersson2025uppercase}
show that meaning-preserving changes (spacing, punctuation, example ordering, case) cause substantial performance shifts.
All sources of non-determinism impact agentic evaluation scores.

Most directly related to our work, \citet{mustahsan2025stochasticity} proposes using intraclass correlation to quantify evaluation stability in agentic systems, showing that stability varies with task complexity and model capability.
\citet{biderman2024lessons} document broader reproducibility challenges in few-shot language model evals, and propose a harness for standardized assessment.
\citet{madaan2024quantifying} propose methods for quantifying and understanding variance in evaluation benchmarks. 
\citet{pimentel2024beyond} conduct an analysis exposing how different evaluation frameworks introduce variability in LLM evals.
\citet{heinemansignal} propose a framework for distinguishing between meaningful signal and noise in evals using a signal-to-noise ratio.
\citet{shen2026sera} apply the same signal-to-noise ratio to assess the reliability of their agentic training findings, and find a median standard deviation of 1.2\% in their experiments with \swebenchverified{}.

Work on agent diversity \citep{audran2025does} demonstrates that behavioral diversity is an important factor in achieving higher performance by enabling creative search of diverse solutions to the same problem.
\citet{wang2022self} propose self-consistency to exploit variance beneficially by sampling multiple reasoning paths.
These works recognize that variance exists in large language model inference, and leverage it to improve performance on a certain task.
Complementing these, our work highlights the importance of accounting for this variance when interpreting agentic evaluation scores. We provide the first extensive analysis of when and why multi-step agent trajectories diverge.

\subsection{Reproducibility in Machine Learning}

Reproducibility challenges are pervasive in science \citep{open2015estimating, baker20161}.
Machine learning research is no exception.
\citet{henderson2018deep} report difficulties in reproducing baselines and \citet{agarwal2021deep} show that the shift to computationally expensive benchmarks led to the detrimental practice of evaluating on a small number of runs per task.
In the large language model domain, similar concerns have led to proposals for standardized reporting with multiple runs and confidence intervals \citep{dodge2019show, biderman2024lessons, miller2024adding}.
Our work extends these methodological insights for LLM prompting to multi-step agentic evals, which also suffer from single or too small number of runs,  misleading researchers and practitioners alike.

\section{Conclusion}

We have demonstrated that randomness fundamentally affects the reliability of agentic evals.
Through \numTrajectories{} trajectories and \numTokens{} tokens across six agentic systems (three models and two scaffolds), we quantified substantial variance in single-run \texttt{pass@1} estimates (2.2--6.0 pp ranges, persisting at temperature 0).
We traced the problem to early trajectory divergence (median within first 1\% of tokens) that cascades through autoregressive conditioning.
We characterized performance envelopes showing gaps up to 24.9 pp between optimistic and pessimistic bounds.
Future work should investigate how dynamic context strategies (e.g., context compactation), widely used in production systems but excluded from our study, affect evaluation variance, and extend this analysis to longer-horizon tasks, where cascading effects may amplify our findings.

\subsubsection*{Acknowledgments}

This work was partially supported by the WASP program funded by Knut and Alice Wallenberg Foundation. Some computation was enabled by resources provided by the National Academic Infrastructure for Supercomputing in Sweden (NAISS).

\bibliography{references}
\bibliographystyle{iclr2026_conference}

\clearpage

\appendix

\section{Statistical Power Analysis for Determining the Number of Runs}
\label{app:power-analysis}

This appendix details the mathematical framework for determining the required number of runs to reliably detect differences in \texttt{pass@1} scores.

Consider two experimental conditions (e.g., two models, two temperatures, or two scaffolds) with true \texttt{pass@1} values $\mu_1$ and $\mu_2$. When we run each condition $n$ times on a benchmark, we obtain sample means $\bar{x}_1$ and $\bar{x}_2$ with standard deviations $\sigma_1$ and $\sigma_2$.

Given a desired significance level $\alpha$ (probability of rejecting $H_0$ when it is true) and statistical power $1-\beta$ (probability of rejecting $H_0$ when it is false), we aim to determine the required number of runs $n$ to reliably detect a difference of magnitude $\Delta = |\mu_1 - \mu_2|$.

We frame this as a two-sample hypothesis test:
\begin{align}
H_0 &: \mu_1 = \mu_2 \quad \text{(no difference)} \\
H_a &: \mu_1 \neq \mu_2 \quad \text{(difference)}
\end{align}

For a two-sample t-test of means with equal sample sizes, assuming known and equal standard deviation $\sigma$, the required number of runs per agent under test ($n_1 = n_2 = n$) can be computed as follows:
\begin{equation}
  t = \frac{\bar{x}_1 - \bar{x}_2}{\sigma \sqrt{\frac{2}{n}}}
\end{equation}

Under the null hypothesis $H_0: \mu_1 = \mu_2$, we reject $H_0$ if $|t| > Z_{\alpha/2}$. Under the alternative hypothesis $H_a: \mu_1 \neq \mu_2$, the expected value of the test statistic is:
\begin{equation}
\mathbb{E}[t] = \frac{\Delta}{\sigma \sqrt{2/n}}
\end{equation}

For the test to achieve power $1-\beta$, the expected test statistic must exceed the critical value by at least $Z_{\beta}$ standard errors:
\begin{equation}
\frac{\Delta}{\sigma \sqrt{2/n}} \geq Z_{\alpha/2} + Z_{\beta}
\end{equation}

\begin{align}
  \sqrt{\frac{n}{2}} &\geq \frac{(Z_{\alpha/2} + Z_{\beta}) \sigma}{\Delta} \\
  n &\geq 2 \left( \frac{Z_{\alpha/2} + Z_{\beta}}{\Delta / \sigma} \right)^2
\end{align}

Where $Z_{\alpha/2}$ is the critical value from the standard normal distribution for a two-tailed test at significance level $\alpha$, and $Z_{\beta}$ is the critical value corresponding to the desired power $1-\beta$.

\begin{figure}[t]
  \centering
  \includegraphics[width=\textwidth]{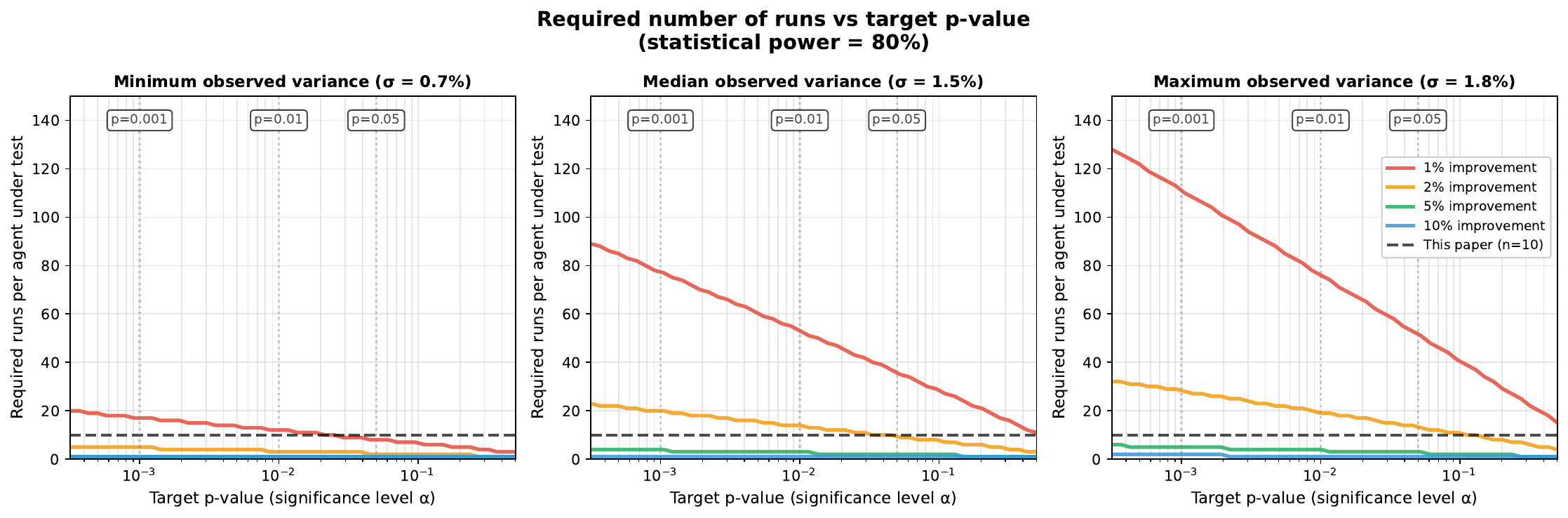}
  \caption{Required number of runs per agent under test for detecting improvements of different magnitudes (1\%, 2\%, 5\%, 10\%) under three variance scenarios observed in our experiments, at significance level $p < 0.05$ and 80\% statistical power. The minimum variance scenario ($\sigma = 0.7\%$) represents the most favorable case, while the maximum variance ($\sigma = 1.8\%$) represents the most challenging evaluation conditions. The exponential increase in required runs for smaller improvements, particularly at higher variance levels, demonstrates that single-run evals cannot reliably distinguish small performance differences from random variations.}
  \label{fig:power-analysis-variance}
\end{figure}

\begin{figure}[t]
\centering
\includegraphics[width=\textwidth]{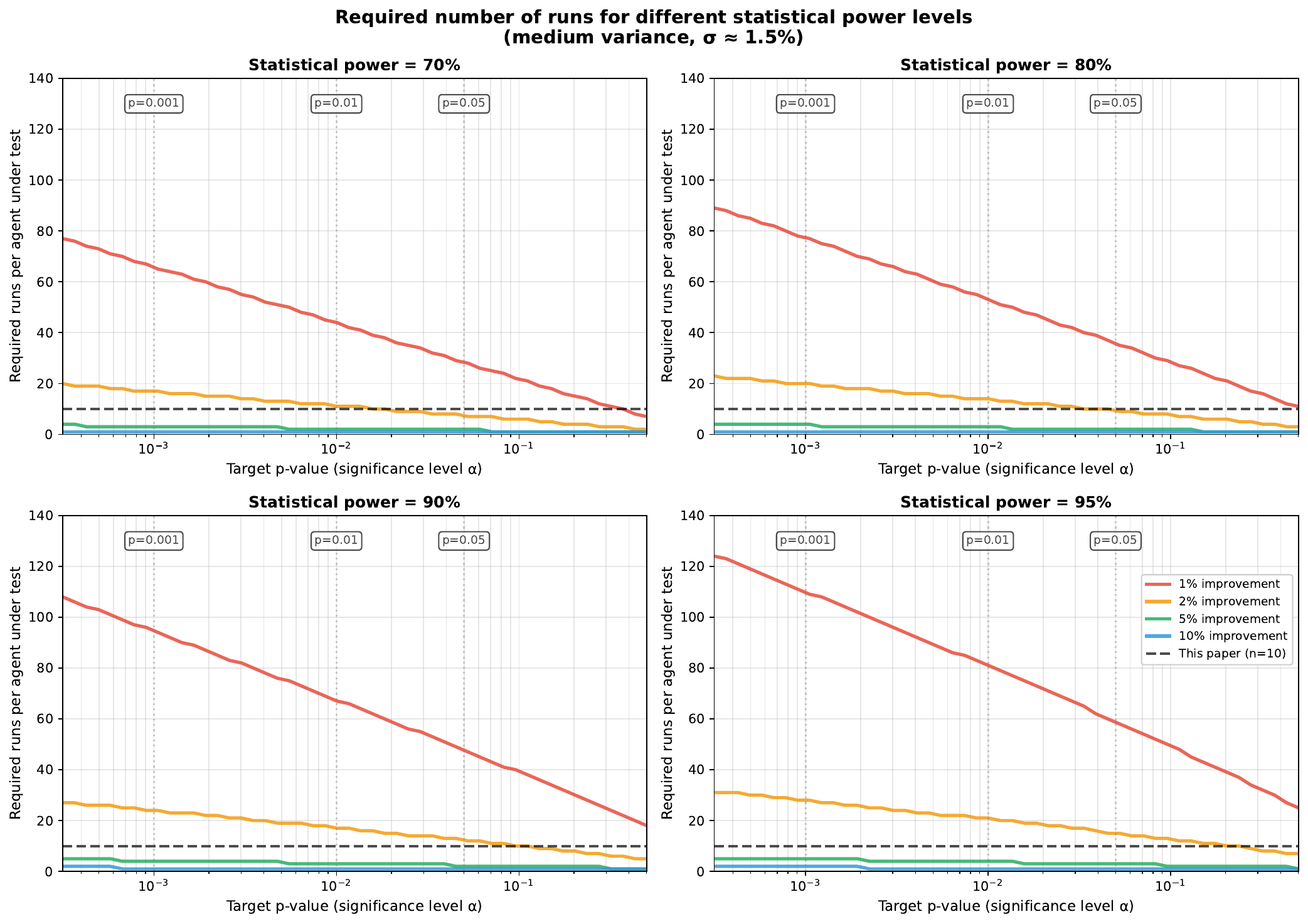}
\caption{Required number of runs per agent under test for detecting improvements of different magnitudes (1\%, 2\%, 5\%, 10\%) at varying statistical power levels (70\%, 80\%, 90\%, 95\%), assuming median observed variance ($\sigma = 1.5\%$) and significance level $p < 0.05$. Higher desired statistical power requires substantially more runs, particularly for detecting small improvements. For example, detecting a 2\% improvement with 80\% power requires 9 runs per agent, while achieving 95\% power for the same effect size requires 15 runs. The exponential growth in required sample size for smaller effect sizes demonstrates why single-run evals are insufficient for reliably detecting small improvements.}
\label{fig:power-analysis-multiple-powers}
\end{figure}

\clearpage

\section{Normality of \texttt{pass@1} Scores}
\label{app:normality}

The power analysis in \Cref{app:power-analysis} assumes that \texttt{pass@1} scores are approximately normally distributed across runs.
We verified this assumption empirically using the Shapiro-Wilk test \citep{shapiro1965analysis}, which tests the null hypothesis that the data are drawn from a normal distribution.

\Cref{tab:normality} reports the results for all 12 configuration--scaffold groups ($n = 10$ runs each).
Eleven of the 12 groups pass the test at $p \geq 0.05$.
The one exception, \texttt{DeepSWE-preview} on \texttt{nano-agent} ($W = 0.80$, $p = 0.016$), is also the group with the smallest observed variance ($\sigma = 1.0\%$).
With $n = 10$, the test has limited power to detect departures from normality, so passing is not a confirmation of exact normality; nonetheless, the results are consistent with the normal approximation being adequate for the purpose of power analysis.

\begin{table}[h]
  \centering
  \caption{Shapiro-Wilk normality test results for \texttt{pass@1} scores across 10 runs per configuration--scaffold group. $W$ is the test statistic; a p-value $\geq 0.05$ indicates no significant departure from normality at the 5\% level.}
  \label{tab:normality}
  \small
  \begin{tabular}{@{}llcccc@{}}
  \toprule
  \textbf{Model} & \textbf{Scaffold} & \textbf{Mean} & \textbf{Std} & $W$ & \textbf{p-value} \\
  \midrule
  \texttt{DeepSWE-preview}       & \texttt{nano-agent} & 31.4\% & 1.0\% & 0.803 & 0.016 \\
  \texttt{DeepSWE-preview}       & \texttt{r2e-gym}    & 34.4\% & 1.5\% & 0.989 & 0.996 \\
  \texttt{DeepSWE-preview-temp0} & \texttt{nano-agent} & 20.4\% & 1.0\% & 0.881 & 0.134 \\
  \texttt{DeepSWE-preview-temp0} & \texttt{r2e-gym}    & 19.2\% & 1.5\% & 0.923 & 0.383 \\
  \texttt{Devstral-2}            & \texttt{nano-agent} & 63.5\% & 1.1\% & 0.894 & 0.190 \\
  \texttt{Devstral-2}            & \texttt{r2e-gym}    & 34.9\% & 1.5\% & 0.956 & 0.733 \\
  \texttt{Devstral-2-temp0}      & \texttt{nano-agent} & 63.8\% & 1.6\% & 0.970 & 0.893 \\
  \texttt{Devstral-2-temp0}      & \texttt{r2e-gym}    & 35.4\% & 1.7\% & 0.952 & 0.695 \\
  \texttt{Qwen3-32B}             & \texttt{nano-agent} & 16.4\% & 0.7\% & 0.885 & 0.149 \\
  \texttt{Qwen3-32B}             & \texttt{r2e-gym}    & 23.9\% & 1.4\% & 0.964 & 0.834 \\
  \texttt{Qwen3-32B-temp0}       & \texttt{nano-agent} & 16.4\% & 1.2\% & 0.962 & 0.809 \\
  \texttt{Qwen3-32B-temp0}       & \texttt{r2e-gym}    & 22.3\% & 1.8\% & 0.964 & 0.830 \\
  \bottomrule
  \end{tabular}
\end{table}

\clearpage

\section{Inference Hyper-Parameters}

This section details the hyperparameters used for each model in our experiments. All locally deployed models (\texttt{Qwen3-32B} and \texttt{DeepSWE-preview}) were hosted using vLLM on NVIDIA A100 80GB GPUs, using a total of approx. 3,500 GPU hours.
\texttt{Devstral-2} was accessed through Mistral's API.

Both scaffolds use their default configurations.
For \texttt{nano-agent}, we set a maximum of 500 tool calls per run, while \texttt{r2e-gym} allows up to 100.

\begin{table}[h]
  \centering
  \caption{Model inference configuration.}
  \label{tab:model-hyperparams}
  \small
  \begin{tabular}{@{}llcccc@{}}
  \toprule
  \textbf{Model} & \textbf{Deployment} & \textbf{Context Limit} & \textbf{Temperature} & \textbf{top\_p} & \textbf{top\_k} \\
  \midrule
  \texttt{Qwen3-32B}       & vLLM (8 x A100 80GB) & 65,536  & 0.6 / 0.0 & 0.95 & 20   \\
  \texttt{DeepSWE-preview} & vLLM (8 x A100 80GB) & 65,536  & 1.0 / 0.0 & ---  & --- \\
  \texttt{Devstral-2}      & Mistral API & 256,000 & 0.2 / 0.0 & ---  & --- \\
  \bottomrule
  \end{tabular}
\end{table}  

\Cref{tab:model-hyperparams} presents the inference hyperparameters for each model.
For \texttt{Qwen3-32B}, we use the sampling parameters used by the authors \citep{yang2025qwen3} when evaluating the model in thinking mode, with the exception of the context limit, which we increase to 65,536 tokens.
For \texttt{DeepSWE-preview}, we use the temperature suggested in the model card.
For \texttt{Devstral-2}, we follow the recommendations in the release post.Temperature 0 experiments use greedy decoding for all models.

\clearpage

\section{Pass@k Plots}

\Cref{fig:pass_at_k} in the main paper shows \texttt{pass@k} and \texttt{pass\^{}k} curves for \texttt{DeepSWE-preview} on both \texttt{nano-agent} and \texttt{r2e-gym}.
For completeness, we provide additional \texttt{pass@k} plots for all other model-scaffold pairs evaluated in this study.

\begin{figure}[t]
\centering
\begin{subfigure}[b]{0.48\textwidth}
\centering
\includegraphics[width=\textwidth]{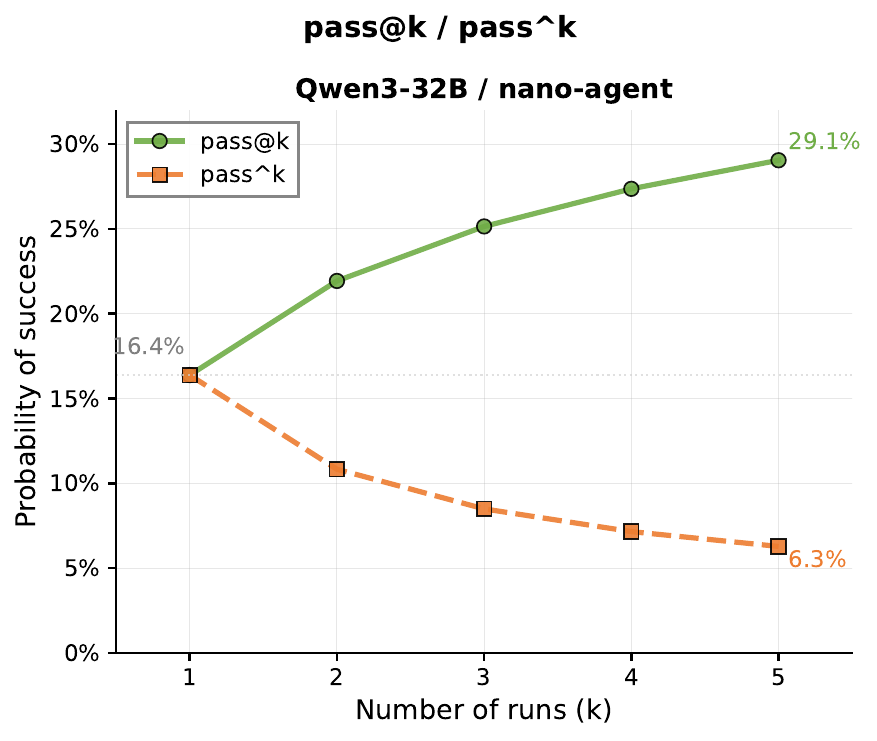}
\end{subfigure}
\hfill
\begin{subfigure}[b]{0.48\textwidth}
\centering
\includegraphics[width=\textwidth]{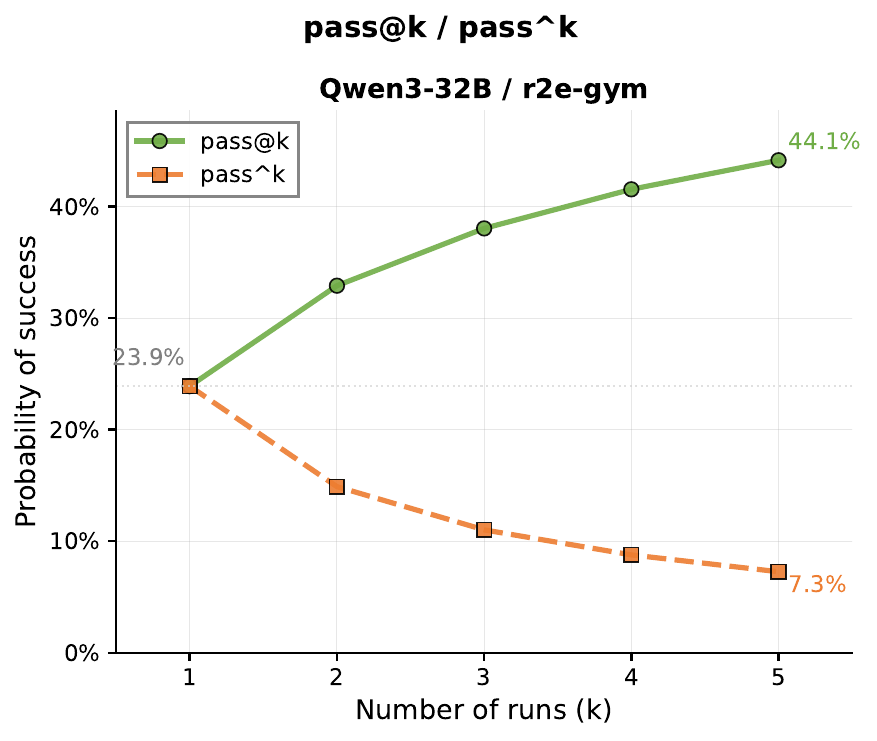}
\end{subfigure}
\vspace{0.5cm}
\begin{subfigure}[b]{0.48\textwidth}
\centering
\includegraphics[width=\textwidth]{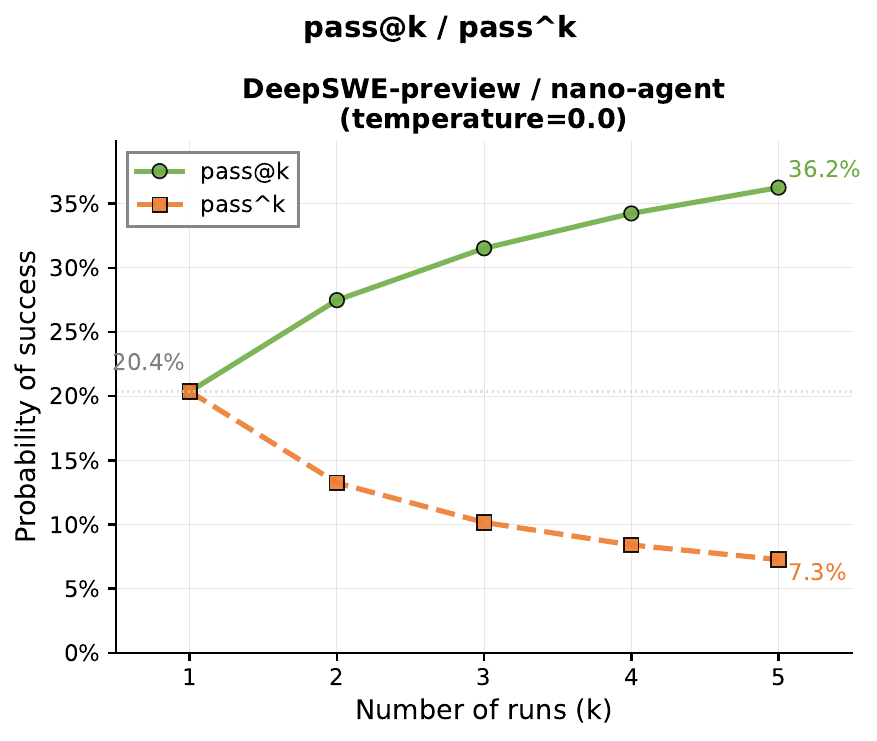}
\end{subfigure}
\hfill
\begin{subfigure}[b]{0.48\textwidth}
\centering
\includegraphics[width=\textwidth]{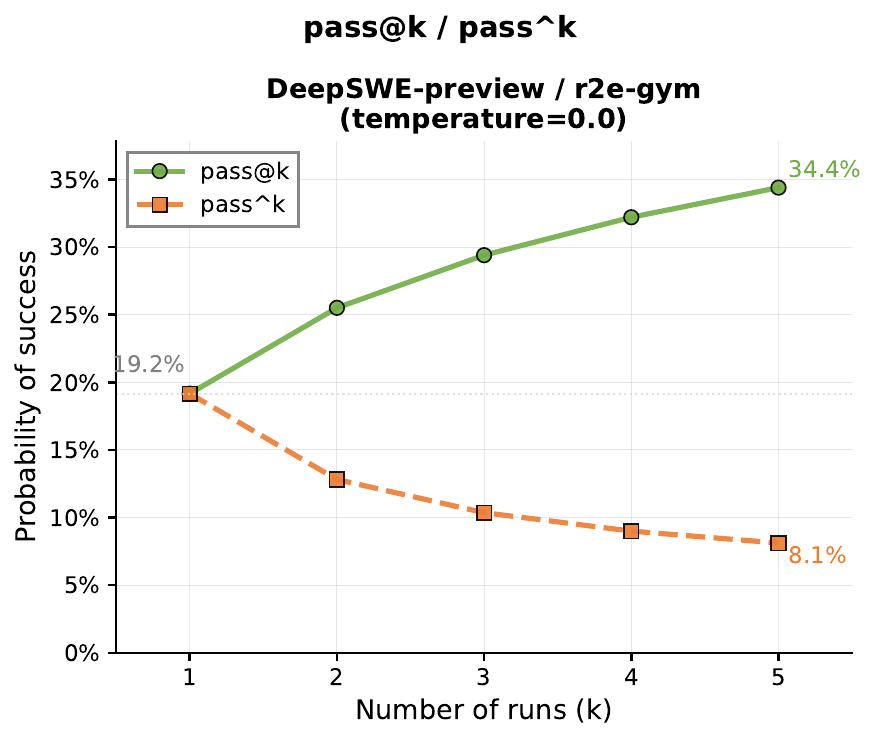}
\end{subfigure}
\vspace{0.5cm}
\begin{subfigure}[b]{0.48\textwidth}
\centering
\includegraphics[width=\textwidth]{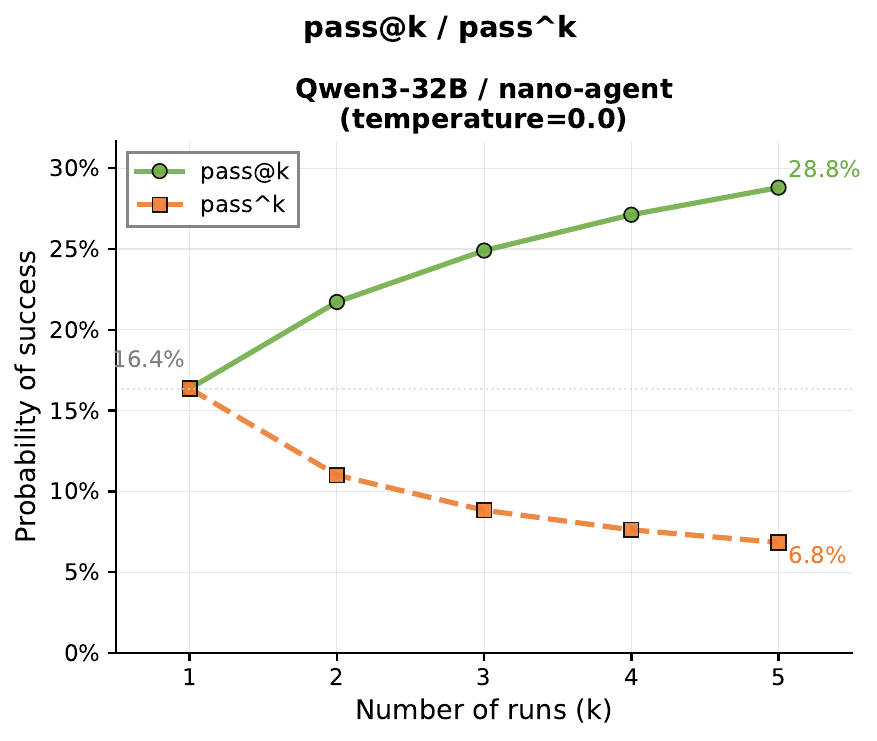}
\end{subfigure}
\hfill
\begin{subfigure}[b]{0.48\textwidth}
\centering
\includegraphics[width=\textwidth]{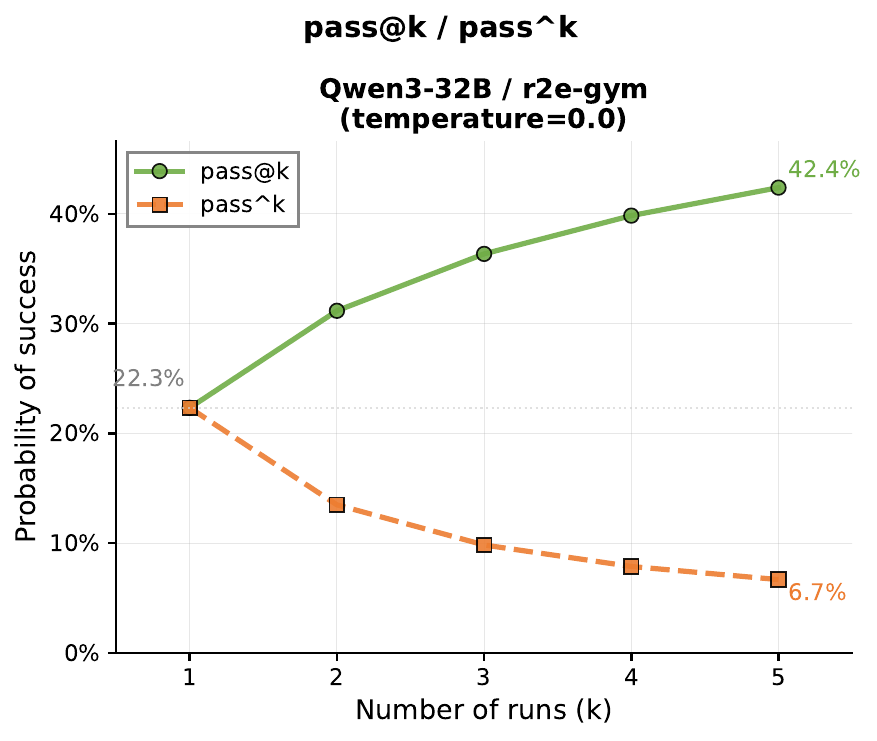}
\caption{\texttt{Qwen3-32B-temp0}\\\texttt{r2e-gym}}
\end{subfigure}
\caption{Additional \texttt{pass@k} and \texttt{pass\^{}k} curves for all model-scaffold pairs (part 1/2).}
\label{fig:pass_at_k_appendix_1}
\end{figure}

\begin{figure}[t]
\centering
\begin{subfigure}[b]{0.48\textwidth}
\centering
\includegraphics[width=\textwidth]{plots/pass_at_k_mistralai_Devstral-2-123B-Instruct-2512_nano-agent.pdf}
\end{subfigure}
\hfill
\begin{subfigure}[b]{0.48\textwidth}
\centering
\includegraphics[width=\textwidth]{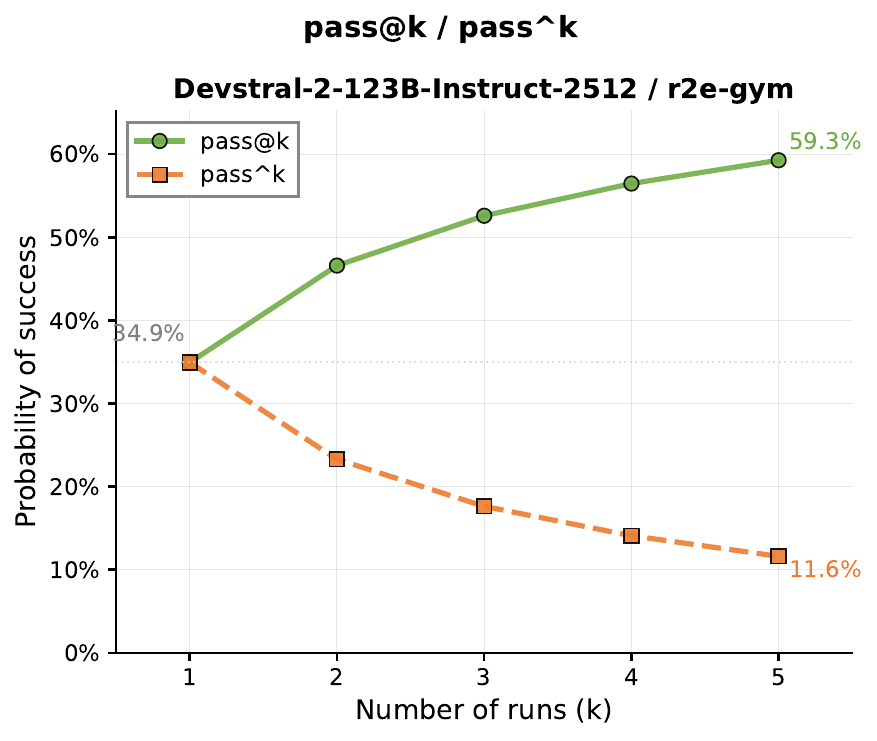}
\end{subfigure}
\vspace{0.5cm}
\begin{subfigure}[b]{0.48\textwidth}
\centering
\includegraphics[width=\textwidth]{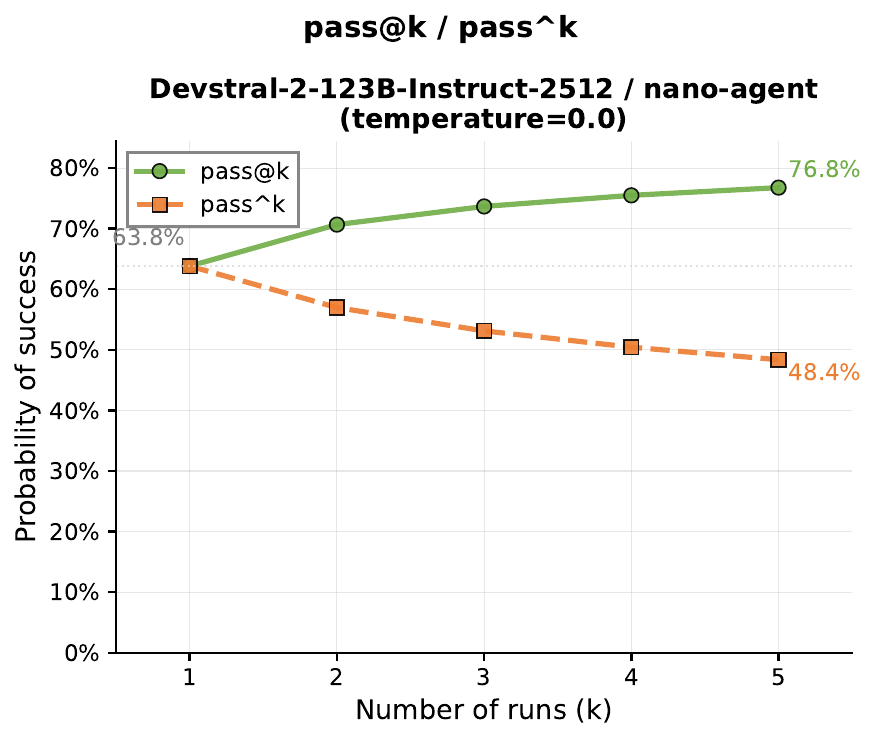}
\end{subfigure}
\hfill
\begin{subfigure}[b]{0.48\textwidth}
\centering
\includegraphics[width=\textwidth]{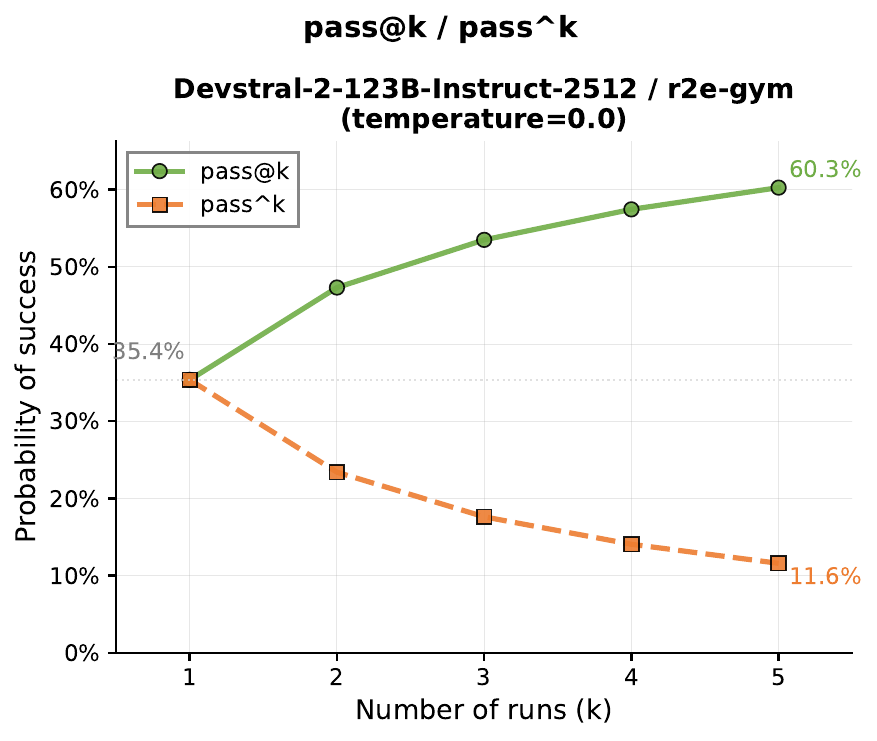}
\end{subfigure}
\caption{Additional \texttt{pass@k} and \texttt{pass\^{}k} curves for all model-scaffold pairs (part 2/2).}
\label{fig:pass_at_k_appendix_2}
\end{figure}

\end{document}